\PassOptionsToPackage{table}{xcolor}
\documentclass[sigconf,screen,nonacm]{acmart}

\usepackage{amsmath}
\usepackage{booktabs}
\usepackage{multirow}
\usepackage{siunitx}

\AtBeginDocument{%
  }

\setcopyright{none}
\renewcommand\footnotetextcopyrightpermission[1]{}

\makeatletter
\def\@Ac@affiliation@missing@error#1{}
\let\@ACM@checkaffil\relax
\makeatother

\newcommand{\compactauthorblock}{%
\begin{tabular}[t]{@{}c@{}}
Qiwei Yan\textsuperscript{1,2,$\dagger$} \quad
Zhiqiang Yuan\textsuperscript{1,$\dagger$,$\S$} \quad
Chongyang Li\textsuperscript{2} \quad
Jiapei Zhang\textsuperscript{1}\\[3pt]
Ying Deng\textsuperscript{1} \quad
Jinchao Zhang\textsuperscript{1,*} \quad
Jie Zhou\textsuperscript{1}
\end{tabular}\\[2pt]
\textsuperscript{1}WeChat AI, Tencent, Beijing, China \quad
\textsuperscript{2}University of Chinese Academy of Sciences, Beijing, China\\[1pt]
\texttt{\small yanqiwei22@mails.ucas.edu.cn, yuanzhiqiang19@mails.ucas.ac.cn}%
}

\author{Qiwei Yan, Zhiqiang Yuan, Chongyang Li, Jiapei Zhang, Ying Deng, Jinchao Zhang, Jie Zhou}
\affiliation{}  % dummy to pass acmart validation
\renewcommand{\shortauthors}{Yan et al.}

\makeatletter
\let\orig@typeset@author@bx\@typeset@author@bx
\def\@typeset@author@bx{%
  \bgroup
  \hsize=\author@bx@wd
  \global\setbox\author@bx=\vtop{%
    \if@ACM@sigchiamode\else\centering\fi
    \@authorfont\compactauthorblock\par
  }%
  \box\author@bx\hspace{\author@bx@sep}%
  \egroup
}
\makeatother

\begin{document}

\title{RAVA: Retrieval-Augmented Viewpoint Alignment for Subject-Driven Image Generation}

% ===== arXiv author information template =====
% Replace the placeholders below with the real author list before uploading.
% ACM's acmart class works best when each author is declared separately.
%
% Optional notes:
% - Equal contribution: add \authornote{Equal contribution.} after the
%   first relevant \author, then add \authornotemark[1] after later
%   authors sharing that note.
% - Corresponding author: add \authornote{Corresponding author.} after
%   the relevant \author, or put the note in the acknowledgments.
% - If an author has multiple affiliations, repeat \affiliation blocks
%   under the same \author.

\begin{abstract}
Reference-driven image generation has made rapid progress on identity preservation, but reliable viewpoint control across different subjects remains poorly understood. The difficulty is not merely generating a new image of the target subject: the model must infer the implicit viewpoint of one subject and transfer it to another subject using only image-level evidence, without camera poses, depth, or ray-based conditions. In this setting, existing generators conditioned on multiple image references often rely on spurious semantic correlations, which lead to viewpoint drift, part-level structural mismatches, and missing or unsupported target-specific content. We formulate this challenge as cross-subject viewpoint alignment and propose RAVA, a retrieval-augmented framework that supplies explicit geometric evidence before generation. RAVA first learns a cross-instance viewpoint embedding that retrieves target-subject images aligned with the anchor viewpoint, then applies a LogDet-based subset selection strategy to retain a compact reference set that is both view-consistent and structurally complementary. The selected references are finally consumed by a fine-tuned multi-reference image generator. Experiments show that generic semantic embeddings are nearly random for this task, while the proposed retriever substantially improves viewpoint retrieval quality. On cross-subject generation, RAVA consistently outperforms zero-shot baselines and stronger retrieval alternatives under the same generation backbone. These results indicate that cross-subject viewpoint alignment benefits from retrieval-augmented geometric grounding rather than relying on end-to-end generation alone.
\end{abstract}




% \begingroup
% \renewcommand{\thefootnote}{\fnsymbol{footnote}}
% \footnotetext[2]{Equal contribution. $^{\S}$ Tech Lead. * Corresponding authors. \\ This work was done during Qiwei Yan's internship at WeChat AI, Tencent Inc. under the guidance of Zhiqiang Yuan.}
% % \footnotetext[3]{This work was done during Kailin Lyu's internship at WeChat AI, Tencent Inc.}
% \endgroup

\maketitle
\renewcommand{\thefootnote}{\fnsymbol{footnote}}
\footnotetext[2]{Equal contribution. $\S$ Tech Lead. $*$ Corresponding author. \\ This work was done during Qiwei Yan's internship at WeChat AI, Tencent Inc. under the guidance of Zhiqiang Yuan.}

\section{Introduction}

\begin{figure}[t]
\centering
\includegraphics[width=\columnwidth]{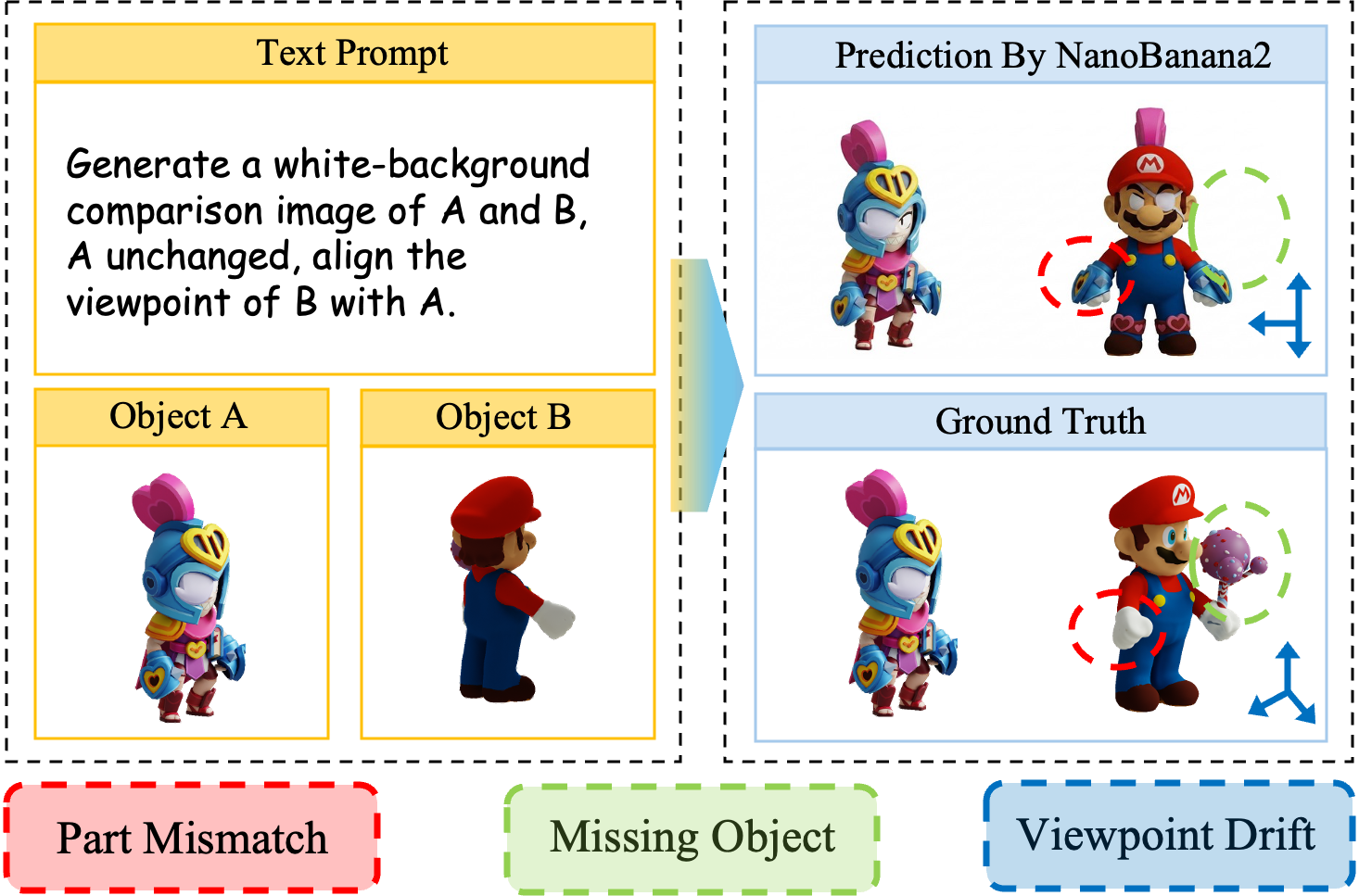}
\caption{Motivating example with NanoBanana2 under cross-subject viewpoint alignment. Even this strong proprietary model exhibits viewpoint drift, part mismatch, and missing target-specific content.}
\Description{A teaser example showing subject A, subject B, a NanoBanana2 prediction, and the ground truth. The prediction exhibits viewpoint drift, part mismatch, and missing target-specific content relative to the desired aligned target view.}
\label{fig:teaser}
\end{figure}

Reference-driven image generation has become increasingly capable of preserving subject identity, combining multiple image prompts, and following open-ended textual instructions. What remains much less reliable is \emph{geometric controllability}: when a user provides one subject as the viewpoint reference and asks the model to render a different subject from the same view, current systems often preserve appearance while failing to preserve geometry. The resulting outputs exhibit recurrent but still insufficiently studied errors, including viewpoint drift, part mismatch, and missing or unsupported content. These failures are especially problematic for multimedia applications that require faithful cross-subject visual comparison.

This paper studies a setting that lies between subject-driven generation and view-conditioned synthesis. We are given an anchor image of subject $A$ and a small candidate pool of subject $B$ captured from different viewpoints. The task is to generate subject $B$ under the viewpoint implied by subject $A$, while keeping the anchor image unchanged and relying only on image-level observations. In our benchmark, this generation target is instantiated as a side-by-side comparison image so that viewpoint fidelity can be evaluated directly. Unlike conventional multi-view synthesis, the input and output do not share instance identity. Unlike camera-conditioned rendering, no explicit camera pose, ray map, depth, or 3D asset is available at inference time. The model therefore has to infer viewpoint from appearance alone and transfer it across subjects with different geometry, texture, and category-specific structure.

This setting exposes a limitation of current pipelines. Multi-subject generation methods are designed primarily around identity preservation, subject disentanglement, and layout control. Multi-view generation methods, in contrast, depend on within-instance consistency and often assume explicit camera control. Simply fine-tuning a stronger multi-image generator does not resolve the mismatch between these assumptions and our task. When the conditioning set provides insufficient geometric evidence, the generator is forced to infer the target view from weak semantic correlations, which is precisely when part-level structural inconsistencies and unsupported content emerge.

Figure~\ref{fig:teaser} is intentionally a \emph{motivation} figure rather than a formal benchmark comparison. We use NanoBanana2 precisely because it is a strong proprietary model: its failure illustrates that the difficulty is not confined to weak generators. Because closed-source systems do not expose a reproducible multi-reference training and evaluation interface, the quantitative study later in the paper focuses on open and controlled baselines under a shared protocol. The broader lesson of Fig.~\ref{fig:teaser} is that the bottleneck is not only the generator, but also the \emph{evidence supplied to the generator}. If the candidate pool of subject $B$ already contains images near the desired view, the central problem becomes how to retrieve and package that evidence before denoising begins. This observation motivates RAVA, a retrieval-augmented view alignment framework that decomposes the task into two preconditioning stages before generation.

This decomposition is effective only if the retriever measures \emph{viewpoint} rather than semantics. Off-the-shelf vision-language embeddings fail here because they are optimized to cluster images by category or appearance, not by cross-instance viewpoint compatibility. We therefore train a dedicated viewpoint embedding with multi-granularity pooling, adaptive gated fusion, and list-level supervision over cross-object candidate groups. We further observe that naive Top-K retrieval wastes the reference budget on near-duplicate views, motivating a LogDet-based reference selection strategy that explicitly encourages complementarity under a small budget.

Experiments support the central claim of the paper: cross-subject viewpoint alignment benefits from explicit geometric grounding before generation. Generic semantic embeddings perform poorly on this retrieval task, while the proposed retriever substantially improves cross-instance viewpoint ranking. On downstream generation, RAVA consistently outperforms zero-shot baselines and stronger retrieval alternatives built on the same fine-tuned Flux.2 backbone. These gains indicate that the value of retrieval in this setting is not generic semantic augmentation, but reliable geometric conditioning.

Our contributions are threefold:
\begin{itemize}
    \item We formalize \emph{cross-subject viewpoint alignment} as a distinct reference-driven generation problem and identify insufficient geometric evidence as a main cause of failure for existing multi-image generators.
    \item We propose a cross-instance viewpoint retrieval module that combines multi-granularity visual-token pooling, adaptive gated fusion, and group-wise list-level supervision to learn viewpoint comparability beyond semantic similarity.
    \item We introduce a LogDet-based reference selection strategy that balances viewpoint fidelity and information complementarity under a tight reference budget and improves downstream generation quality.
\end{itemize}

\section{Related Work}

\noindent\textbf{Multi-subject Driven Generation.} Subject-driven and multi-subject generation methods focus on preserving user-specified identities under diverse prompts. Early personalization methods learn subject tokens or fine-tune diffusion backbones from a few exemplars \cite{gal2022image,ruiz2023dreambooth}, while subsequent works improve subject representation, editing ability, and data efficiency \cite{li2023blip,li2023dreamedit,hua2023dreamtuner,chen2023subject}. Recent systems also reduce training cost, strengthen prompt-following, or scale to multiple image references and multiple identities \cite{chan2024improving,kang2025flux,wu2025less,miao2024subject,he2025anystory,kumari2023multi,ding2024freecustom,jiang2025mc}. In multi-subject settings, much of the literature focuses on preventing subject leakage and enforcing cleaner composition through attention isolation, causal training, or layout constraints \cite{dahary2024yourself,li2025customized,tran2026disenid,peng2025muse,dahary2025decisive}. Our setting differs in emphasis: the main challenge is not identity composition itself, but accurate transfer of viewpoint from one subject to another under image-only conditioning.

\noindent\textbf{Multi-view Image Generation.} Multi-view synthesis methods generate novel views of the same object or scene by exploiting within-instance geometric consistency. Representative directions include 3D-aware generators \cite{zhang2022multi,zhao2018multi}, single-image diffusion priors for novel view synthesis \cite{shi2023zero123++}, synchronized or jointly sampled multi-view diffusion \cite{liu2023syncdreamer,shi2023mvdream,di2024dimvis}, and geometry-enhanced models that inject depth, camera control, epipolar constraints, or ray-based features \cite{yang2024consistnet,hollein2024viewdiff,voleti2024sv3d,huang2025mv,kang2025multi}. Multi-reference variants further explore unposed fusion, consistency-preserving denoising, and depth-aware conditioning \cite{sun2018multi,yang2024viewfusion,kani2023upfusion,hu2024mvd,xu2025flexgen}. These methods are powerful when explicit geometry is available or when all views correspond to the \emph{same} instance. By contrast, our task requires cross-instance viewpoint transfer without camera supervision.

\noindent\textbf{Viewpoint-aware Representation Learning.} Generic visual and vision-language embeddings are effective for semantic retrieval, but they are not designed to preserve viewpoint comparability across different objects. This has motivated a line of work on viewpoint- or pose-sensitive representations. Early works exploit rendered multi-view supervision or latent viewpoint alignment to learn view-discriminative descriptors \cite{su2015rendercnn,kanezaki2018rotationnet,poier2018pose}, while more recent methods introduce explicit viewpoint encodings or reference-based pipelines for object pose estimation \cite{cai2022ove6d,liu2025unopose}. However, these methods mainly target explicit pose prediction, same-instance alignment, or geometry-aware matching, often with rendered, depth, or RGB-D inputs. By contrast, our setting requires cross-instance viewpoint comparability under image-only conditioning, and the learned embedding serves downstream image generation rather than explicit pose recovery.

\noindent\textbf{Retrieval-Augmented Image Generation.} Retrieval-augmented generation supplies external evidence to image synthesis, alleviating the burden on model parameters alone. Prior work retrieves nearest neighbors for diffusion guidance \cite{blattmann2022retrieval,sheynin2022knn,rombach2022text}, grounds rare concepts or long-tail prompts with retrieved multimodal evidence \cite{chen2022re,shalev2025imagerag,yuan2025finerag,zhu2025cross}, and improves realism or factuality through reflective or uncertainty-aware retrieval \cite{lyu2025realrag,li2025ia,qi2025ar,cioni2023diffusion,wan2024factuality,zhang2024garmentaligner}. However, most retrieval criteria are semantic: retrieved images are relevant in category, style, or prompt, but not necessarily in geometry. Our work positions retrieval as a \emph{view alignment} mechanism, where the core challenge is selecting references that provide compatible and non-redundant viewpoint evidence.

\section{Method}

\subsection{Problem Formulation}

\begin{figure*}[t]
\centering
\includegraphics[width=\textwidth]{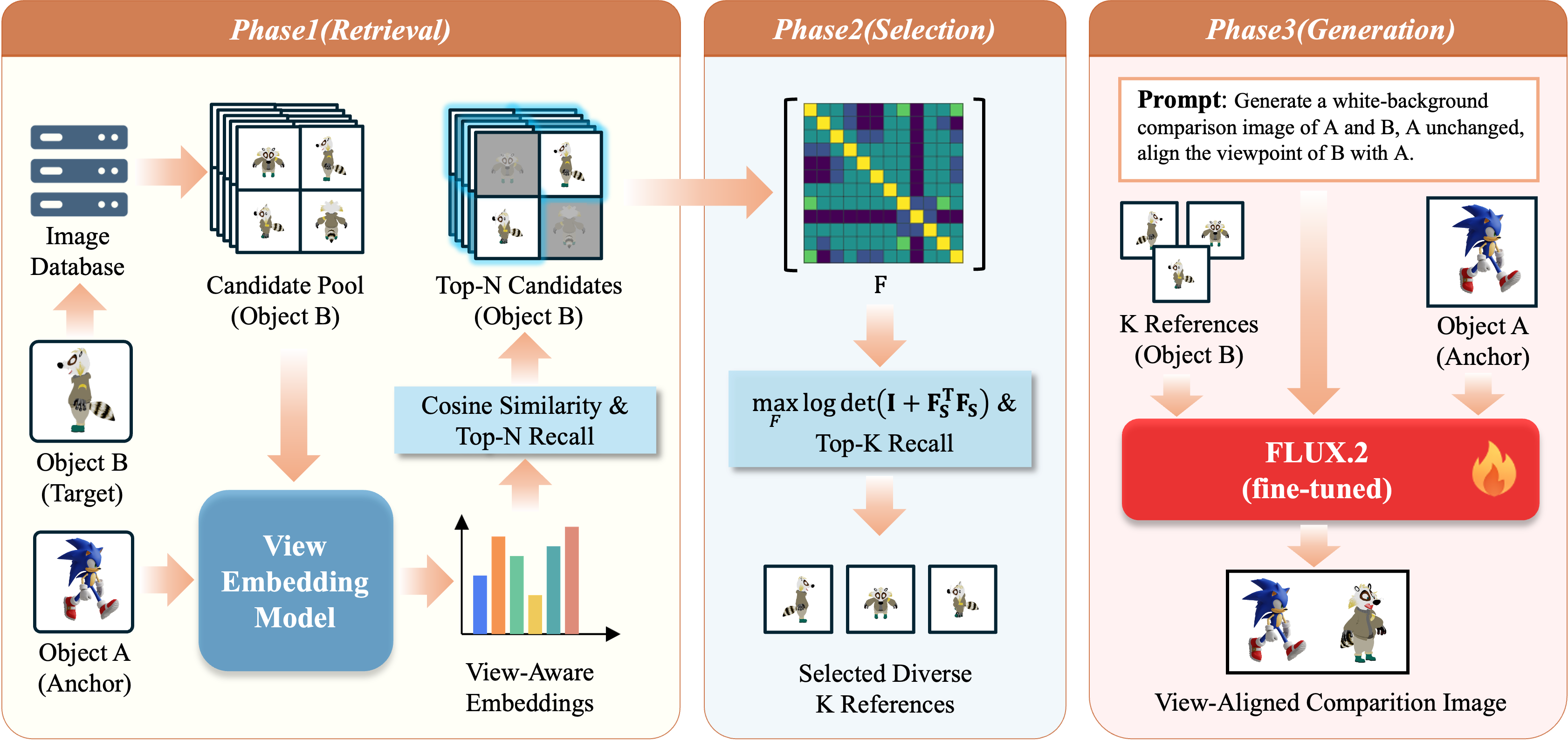}
\caption{Overview of RAVA. \textbf{Phase~1}: a cross-instance viewpoint embedding recalls the top-$M$ target-subject candidates most compatible with the anchor viewpoint. \textbf{Phase~2}: a LogDet objective selects $K$ aligned yet complementary references. \textbf{Phase~3}: the selected references and the anchor image are fed into a fine-tuned Flux.2 generator.}
\Description{A three-phase pipeline diagram showing the RAVA framework. The first phase retrieves target-subject candidate images using a cross-instance viewpoint embedding, the second phase selects a compact complementary subset with a LogDet objective, and the third phase feeds the selected references and anchor image into a fine-tuned Flux.2 generator.}
\label{fig:framework}
\end{figure*}

We study a reference-driven generation problem that requires \emph{cross-subject viewpoint alignment} under minimal geometric supervision. The input contains an anchor image $x_A$ of subject $A$ captured from an unknown viewpoint, together with a candidate pool $\mathcal{C}_B = \{x_B^1, \ldots, x_B^N\}$ of a different subject $B$ observed from multiple views. The objective is to synthesize an image $\hat{x}_B$ in which subject $B$ is rendered from the viewpoint implied by $x_A$:
\begin{equation}
\hat{x}_B = G(x_A, \mathcal{C}_B).
\end{equation}
In our implementation, the model outputs a side-by-side comparison canvas that contains the preserved anchor view of $A$ and the generated target view of $B$.
No camera parameters, depth, ray maps, or 3D models are available for either subject. The viewpoint is therefore specified only implicitly through visual appearance, while the candidate pool may contain substantial variation in orientation, shape, and texture.

This setting differs from both conventional multi-view synthesis and camera-conditioned generation. In multi-view synthesis, all views typically correspond to the same object or scene, which makes correspondence easier to establish. In camera-conditioned generation, the target pose is provided explicitly. It also differs from 3DGS-style reconstruction, where the goal is to recover an explicit 3D scene or object representation before rendering novel views. Our task provides neither advantage: the model must infer the anchor viewpoint and match it to a different subject under image-only conditioning. Empirically, this is where end-to-end generators exhibit viewpoint drift, part-level structural mismatches, and missing or unsupported target-specific content.

RAVA addresses this failure mode by decomposing the problem into evidence acquisition and image generation. Rather than passing the full candidate pool to the generator, we first retrieve views of subject $B$ that are compatible with the anchor viewpoint, then compress them into a small but informative subset before generation.

\subsection{Framework Overview}

Figure~\ref{fig:framework} summarizes the pipeline. The method is motivated by a central observation: under image-only conditioning, the desired target view is latent, so direct generation is underdetermined unless the reference set already carries sufficient geometric evidence. We therefore cast RAVA as an \emph{evidence acquisition} problem under a small conditioning budget. Before denoising begins, the framework first estimates which target-subject images are most compatible with the anchor viewpoint and then compresses the recalled pool into a compact set that preserves a reliable viewpoint anchor while covering complementary evidence. This yields the following three computational stages:
\begin{enumerate}
    \item \textbf{View-aware retrieval.} We learn an embedding function $f(\cdot)$ that makes images of \emph{different} objects comparable by viewpoint. Given $x_A$ and each candidate $x_B^i$, we compute a viewpoint similarity score $\mathrm{ViewSim}(x_A, x_B^i)$ and retain the top-$M$ candidates.
    \item \textbf{Budget-aware reference selection.} From the recalled set, we choose a subset of size $K$ that preserves one strong viewpoint anchor while maximizing complementarity among the remaining references through a LogDet criterion.
    \item \textbf{Conditional generation.} The selected references and the anchor image are fed to a fine-tuned multi-image generator to synthesize $\hat{x}_B$.
\end{enumerate}
The key design principle is that the generator should receive a conditioning set that is already geometrically meaningful. RAVA is therefore largely agnostic to the exact generative backbone and is best understood as a geometry-oriented conditioning interface rather than as a modification of the denoising architecture itself.

\subsection{Cross-instance Viewpoint Embedding for View-aware Retrieval}
\label{sec:view_embedding}

The retrieval module aims to learn an embedding function $f(\cdot)$ whose induced similarity is \emph{order-preserving} with respect to cross-instance viewpoint distance. Given an anchor $x_A=I(O_A,v_A)$ and two candidates $x_i=I(O_i,v_i)$ and $x_j=I(O_j,v_j)$, the desired ranking should favor $x_i$ over $x_j$ whenever $v_i$ is closer to $v_A$ than $v_j$, irrespective of object identity. This requirement is stricter than generic semantic retrieval and different from explicit pose estimation: the representation must suppress category- and texture-specific variation while retaining the geometric cues that make different objects comparable by viewpoint under RGB-only conditioning.

Unlike prior pose-sensitive representations that are often optimized for explicit pose recovery or same-instance alignment, our retriever must also remain useful for downstream generation. Standard vision-language embeddings are poorly aligned with this objective because they primarily organize images by category, semantics, or appearance. As a result, two objects viewed from similar directions are often embedded farther apart than two views of the same object seen from different directions.

We model a viewpoint as a viewing direction $v \in \mathbb{S}^{2}$ and denote by $I(O,v)$ the image of object $O$ rendered under direction $v$. The desired behavior of the embedding is monotonic with respect to cross-instance viewpoint distance: for distinct objects $O_1$ and $O_2$, views with smaller angular distance between viewing directions should receive higher similarity than views with larger distance. We instantiate viewpoint similarity by a shifted cosine similarity between normalized embeddings,
\begin{equation}
\mathrm{ViewSim}(x_A, x_i) = \frac{1 + f(x_A)^\top f(x_i)}{2},
\end{equation}
which maps the score to $[0,1]$ and keeps the quality weights in Sec.~\ref{sec:logdet} well defined. We optimize the embedding so that this similarity induces the correct ranking over candidate views.

\noindent\textbf{Backbone and visual tokens.}
We build the retriever on top of the pre-trained Qwen3-VL model. To bias the backbone toward viewpoint-sensitive cues, we pair each image with the fixed prompt \textit{``Represent the viewing angle of this 3D object image.''} Given an input image $x$, the backbone produces a last-layer hidden-state sequence $H=[h_1,\ldots,h_T]\in\mathbb{R}^{T\times d}$ that mixes textual and visual tokens. We isolate the visual-token subset $V\in\mathbb{R}^{N\times d}$ using the model's vision delimiters and discard text tokens, which would otherwise leak semantics into the viewpoint representation. The backbone itself is frozen so that training concentrates on learning a viewpoint-sensitive aggregation head rather than overwriting the broad visual priors of the foundation model.

\noindent\textbf{Multi-granularity pooling.}
Viewpoint evidence is heterogeneous across objects. For some shapes, coarse silhouette already determines the viewing direction, whereas symmetric or structurally complex objects are disambiguated only by asymmetric parts or region-level cues. A single pooling operator is therefore statistically brittle for cross-object viewpoint retrieval. We instead aggregate the visual token sequence at two complementary granularities:
\begin{itemize}
    \item \emph{Global pooling}: we extract a sequence-level descriptor $g\in\mathbb{R}^d$ from the last valid visual token, encoding coarse pose and overall context.
    \item \emph{Regional pooling}: we use a learnable query vector $q_{\mathrm{reg}}\in\mathbb{R}^d$ to perform multi-head attention over the visual token sequence. This produces a region-aware descriptor $r=\mathrm{MHA}(q_{\mathrm{reg}}, V, V)\in\mathbb{R}^d$ that focuses on geometrically discriminative regions.
\end{itemize}

\noindent\textbf{Gated fusion.}
The usefulness of the two descriptors varies across inputs. Fixed-weight fusion may overtrust global silhouette when symmetry makes pose ambiguous, or overemphasize regional evidence when the object can already be resolved reliably from the full contour. We therefore use adaptive gated fusion, which can be interpreted as an input-dependent reliability weighting over two geometric evidence sources. The model first predicts normalized fusion weights
\begin{equation}
\alpha = \mathrm{softmax}(W\,[g;r]) \in \mathbb{R}^{2},
\end{equation}
where $W$ is a learnable projection. The fused representation is then computed as
\begin{equation}
z = \phi\left(\alpha_1 g + \alpha_2 r\right),
\qquad
f(x)=\frac{z}{\|z\|_2},
\end{equation}
where $\phi$ is a lightweight two-layer MLP. The gating network thus performs a data-driven allocation of representational capacity between coarse pose context and region-specific disambiguation.

\begin{figure*}[t]
\centering
\includegraphics[width=\textwidth]{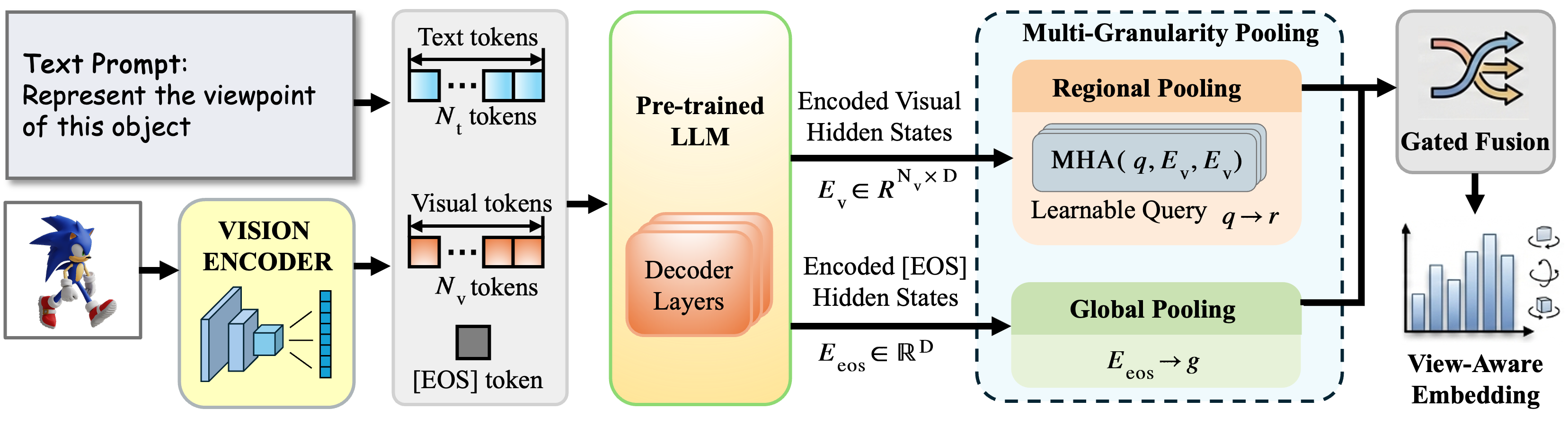}
\caption{Architecture of the cross-instance viewpoint embedding model. Qwen3-VL encodes an input image and a fixed prompt into visual tokens, which are aggregated by regional and global pooling branches and fused by a gated module to produce the final normalized viewpoint embedding.}
\Description{A block diagram showing the cross-instance viewpoint embedding architecture. An input image and a fixed prompt are encoded into visual tokens, which are processed through regional and global pooling branches and then fused by a gated module to produce the final view-aware embedding.}
\label{fig:embedding_model}
\end{figure*}

\noindent\textbf{Training objective.}
Two design choices are important during training. First, supervision is organized \emph{group-wise}: each training instance contains one query image and multiple candidates from different objects. This prevents the model from solving the task by memorizing object appearance and instead forces it to rank heterogeneous candidates by viewpoint. Second, the loss is \emph{list-level} rather than pairwise, because the retriever is consumed only through the ordering it induces over a full candidate pool at inference time.

For each group with candidates $\{x_i\}_{i=1}^n$ and target similarities $\{y_i\}_{i=1}^n$, predicted viewpoint scores are converted to a list distribution
\begin{equation}
p_i = \frac{\exp(\mathrm{ViewSim}(x_A, x_i))}{\sum_{j=1}^{n} \exp(\mathrm{ViewSim}(x_A, x_j))},
\qquad
\tilde{y}_i = \frac{y_i}{\sum_{j=1}^{n} y_j},
\end{equation}
and we optimize
\begin{equation}
\mathcal{L}=\mathrm{KL}(\tilde{y} \| p),
\end{equation}
which directly supervises the ordering over the full candidate list and reduces the train-inference mismatch that would arise from regression or pairwise contrastive losses.

\subsection{View-conditioned Reference Selection via LogDet Maximization}
\label{sec:logdet}

Retrieval alone solves only the \emph{fidelity} part of the problem. Under a tight reference budget, the generator still should not consume every recalled image, because naive Top-K selection tends to return near-duplicate views: alignment is high, but the references add little new information about the unseen geometry of subject $B$. The conditioning set should therefore satisfy two criteria simultaneously: it must contain at least one high-confidence view anchor, and the remaining references should cover complementary evidence rather than repeating the same observation. We formulate this as a constrained subset selection problem,
\begin{equation}
S \subseteq \mathcal{C}_B, \qquad |S| = K,
\end{equation}
where $K$ denotes the reference budget. The objective jointly optimizes viewpoint fidelity and information complementarity.

\noindent\textbf{View-conditioned candidate recall.}
For each candidate image $x_i \in \mathcal{C}_B$, we compute a viewpoint similarity score
\begin{equation}
s_i = \mathrm{ViewSim}(x_A, x_i) \in [0,1],
\end{equation}
using the retriever from Sec.~\ref{sec:view_embedding}. We rank all candidates by $s_i$ and retain the top-$M$ views to form a recall set $\mathcal{R}$ with $M=20$. This separates coarse fidelity filtering from fine-grained subset optimization: clearly incompatible views are removed early, while enough candidates remain for complementary selection inside $\mathcal{R}$.

\noindent\textbf{Anchor view enforcement.}
We decouple viewpoint anchoring from coverage maximization. To ensure that the selected subset contains at least one strong geometric anchor, we always include the highest-scoring candidate
\begin{equation}
x^{\mathrm{anchor}} = \arg\max_{x_i \in \mathcal{R}} s_i
\end{equation}
in the final reference set.

\noindent\textbf{Quality-weighted similarity kernel.}
Within the recalled pool, selection should prefer references that are individually reliable yet collectively non-redundant. We encode these two requirements through a quality-weighted similarity kernel. Each candidate receives a quality weight
\begin{equation}
w_i = s_i^{\gamma},
\end{equation}
where $\gamma \geq 1$ controls how strongly the selector favors near-perfect viewpoint matches. For each recalled image $x_i \in \mathcal{R}$, we reuse the normalized embedding
\begin{equation}
u_i = f(x_i),
\end{equation}
and define a similarity kernel
\begin{equation}
K_{ij} = \exp\!\left(\frac{u_i^\top u_j}{\tau}\right),
\end{equation}
where $\tau$ controls how aggressively redundant candidates are penalized. The corresponding quality-weighted $L$-ensemble matrix is
\begin{equation}
L = \mathrm{diag}(w)\, K\, \mathrm{diag}(w).
\end{equation}
Because the recalled pool has already been filtered by viewpoint similarity, pairwise similarity among the recalled embeddings primarily reflects redundant geometric evidence rather than gross view mismatch. This makes the same embedding space a lightweight but effective proxy for complementarity during subset selection.

\begin{table*}[t]
  \centering
  \caption{Viewpoint retrieval results on the 10k-instance test split, where each query ranks 20 cross-object candidates. Methods marked with $\dagger$ denote task-specific fine-tuning. Best results are in \textbf{bold}.}
  \label{tab:viewpoint_retrieval}
  \setlength{\tabcolsep}{12pt}
  \begin{tabular}{l *{6}{S[table-format=1.3]}}
    \toprule
    \multirow{2}{*}{Method} & \multicolumn{4}{c}{NDCG} & \multicolumn{2}{c}{Correlation} \\
    \cmidrule(lr){2-5} \cmidrule(l){6-7}
    & {@1} & {@3} & {@5} & {@10} & {Concord.} & {Spearman} \\
    \midrule
    Random            & 0.345 & 0.397 & 0.435 & 0.524 & 0.500 & 0.000 \\
    VGGT              & 0.359 & 0.426 & 0.471 & 0.558 & 0.525 & 0.049 \\
    DINOv3            & 0.335 & 0.405 & 0.452 & 0.543 & 0.513 & 0.022 \\
    CLIP              & 0.334 & 0.402 & 0.448 & 0.539 & 0.511 & 0.023 \\
    Qwen3-VL          & 0.329 & 0.396 & 0.442 & 0.533 & 0.505 & 0.011 \\
    \midrule
    VGGT$\dagger$     & 0.404 & 0.463 & 0.506 & 0.594 & 0.565 & 0.121 \\
    DINOv3$\dagger$   & 0.678 & 0.732 & 0.760 & 0.806 & 0.773 & 0.588 \\
    CLIP$\dagger$     & 0.498 & 0.555 & 0.593 & 0.664 & 0.634 & 0.285 \\
    Qwen3-VL$\dagger$ & 0.686 & 0.743 & 0.782 & 0.807 & 0.781 & 0.626 \\
    \midrule
    \rowcolor{gray!10}
    Ours              & {\bfseries 0.750} & {\bfseries 0.797} & {\bfseries 0.822} & {\bfseries 0.859} & {\bfseries 0.832} & {\bfseries 0.710} \\
    \bottomrule
  \end{tabular}
\end{table*}

\noindent\textbf{LogDet-based subset selection.}
Under a Gaussian interpretation of the $L$-ensemble, the log-determinant of a principal submatrix measures the information volume covered by the selected set. Maximizing $\log\det(\cdot)$ therefore implements our two desiderata in a single criterion: high-quality items increase the scale of the volume through $q_i$, while redundant items collapse the volume through the kernel correlations. Conditioned on the mandatory anchor view, we choose the remaining $K-1$ references by solving
\begin{equation}
\max_{T \subseteq \mathcal{R} \setminus \{x^{\mathrm{anchor}}\},\ |T|=K-1}
\ \log\det\!\left(I_{K} + L_{\{x^{\mathrm{anchor}}\} \cup T}\right).
\end{equation}
The final selected subset is
\begin{equation}
S = \{x^{\mathrm{anchor}}\} \cup T^{\ast},
\end{equation}
where $T^{\ast}$ denotes the optimizer. Because the recalled pool and final reference budget are both small in our setting, we solve this objective exactly by enumeration rather than relying on greedy approximation.

\subsection{Generation Backbone}

The final generation stage is deliberately kept fixed so that downstream gains can be attributed to the quality of the conditioning set rather than to architectural changes in the denoiser. Given anchor image $x_A$ and selected reference set $S$, we fine-tune a multi-image Flux.2 backbone to predict the target output under the aligned viewpoint. In our benchmark, the generator receives one anchor image of subject $A$ together with $k$ retrieved reference images of subject $B$, and outputs a side-by-side comparison image that contains the preserved anchor view of $A$ and the generated view of $B$. RAVA does not alter the denoising dynamics or introduce generator-specific modules; its contribution is to reshape the geometric evidence presented to the generator before synthesis begins. All retrieval strategies in Table~\ref{tab:generation} therefore share the same fine-tuned Flux.2 backbone and training recipe.

\begin{table}[t]
\centering
\small
\caption{Cross-subject view-aligned generation results on the 1k-pair test split. \textit{Zero-shot Baselines} use one anchor image of subject $A$ and one reference image of subject $B$. \textit{Retrieval Strategies} share the same fine-tuned Flux.2 backbone and use one anchor image of subject $A$ together with $K=3$ selected reference images of subject $B$. Best results are in \textbf{bold}.}
\label{tab:generation}
\setlength{\tabcolsep}{8pt}
\begin{tabular}{clccc}
\toprule
 & Method & PSNR $\uparrow$ & SSIM $\uparrow$ & LPIPS $\downarrow$ \\
\midrule
\multirow{5}{*}{\shortstack[c]{\textit{Zero-shot}\\\textit{Baselines}}}
 & Qwen-Image     & 10.00 & 0.6915 & 0.4726 \\
 & Flux.2         & 13.22 & 0.7920 & 0.3042 \\
 & UNO            &  9.79 & 0.6866 & 0.5435 \\
 & MS-Diffusion   &  7.26 & 0.6214 & 0.7869 \\
 & OmniGen        &  8.03 & 0.5756 & 0.6167 \\
\midrule
\multirow{5}{*}{\shortstack[c]{\textit{Retrieval}\\\textit{Strategies}\\\footnotesize{(Flux.2-FT)}}}
 & Random         & 15.01 & 0.8249 & 0.2143 \\
 & CLIP           & 15.17 & 0.8278 & 0.2091 \\
 & DINOv3         & 15.03 & 0.8282 & 0.2104 \\
 & Top-K          & 15.40 & 0.8308 & 0.2009 \\
 \rowcolor{gray!10}
 & RAVA (Ours)    & \textbf{15.80} & \textbf{0.8398} & \textbf{0.1829} \\
\bottomrule
\end{tabular}
\end{table}

\section{Experiment}

\subsection{Experimental Setup}

\noindent\textbf{Cross-instance viewpoint dataset.}
We build the benchmark from Objaverse-XL assets. A VLM-based filter retains objects whose front/back and left/right views are visually distinguishable, after which object identities are split into 20k training objects and 2k test objects. Each object is rendered in Blender at 24 anchor viewpoints from a $3\times 8$ elevation-azimuth grid plus random perturbations, yielding 48 views per object.

For viewpoint retrieval, each training sample pairs one query object with a different candidate object. We randomly draw 6 query images; for each query, we sample 2 positives ($<45^\circ$), 2 hard negatives ($45^\circ$--$90^\circ$), and 2 easy negatives ($\geq 90^\circ$), producing about 120k training instances. Given view direction vectors $v_q$ and $v_c$, we compute the angular difference
\begin{equation}
\theta=\arccos(v_q^\top v_c),
\end{equation}
and convert it into a soft viewpoint-similarity target with spherical Gaussian bandwidth $\sigma=45^\circ$:
\begin{equation}
y=\exp\left(-\frac{\theta^2}{2\sigma^2}\right)\in[0,1].
\end{equation}
The retrieval test set follows a 1-query-vs.-20-candidates protocol with 10k instances.

\begin{figure*}[t]
\centering
\includegraphics[width=\textwidth]{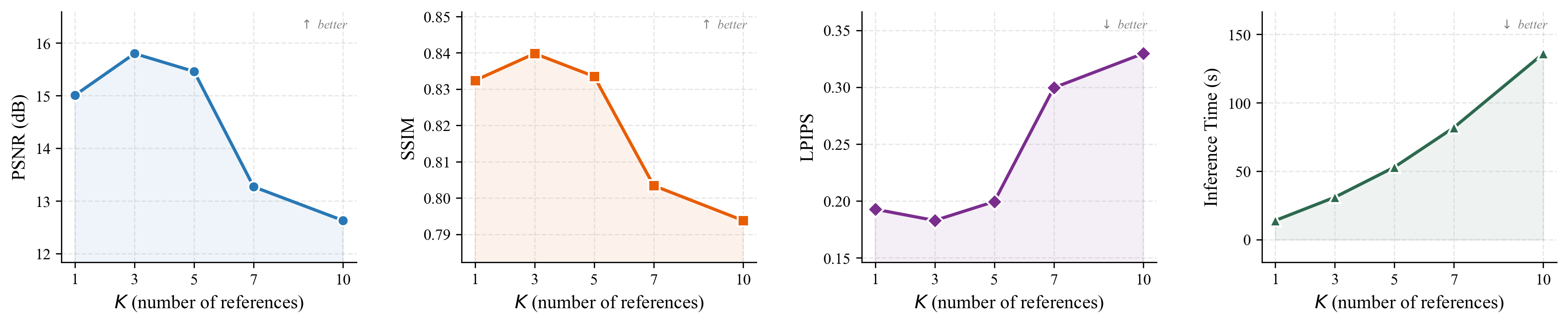}
\caption{Effect of reference budget $K$ on generation quality and inference time. Quality peaks at $K=3$, while inference time increases monotonically.}
\Description{A line chart showing PSNR, SSIM, LPIPS, and inference time as functions of the reference budget K. Generation quality peaks around K equals 3 and degrades for larger K, while inference time increases linearly.}
\label{fig:ablation_k}
\end{figure*}

For downstream generation, we use the same object split and construct 30k training pairs and 1k test pairs. For each pair $(A,B)$, we randomly choose one shared anchor viewpoint as the target view, use the rendering of $A$ at that view as the anchor input, and provide subject $B$ with 1--10 random anchor-view images during training and one image during testing. The supervision target is a white-background side-by-side rendering of $A$ and $B$ at the target viewpoint with resolution $1024\times 512$, and the prompt asks the model to preserve the appearance of $A$ while aligning the viewpoint of $B$ to that of $A$. At test time, the model therefore receives one anchor image of subject $A$ together with a small set of reference images of subject $B$, and outputs a side-by-side comparison image.

\noindent\textbf{Evaluation metrics.}
For viewpoint retrieval, we report NDCG@K, Pairwise Concordance, and Spearman's $\rho$. For generation, we report PSNR, SSIM, and LPIPS against the ground-truth side-by-side target image. Unlike DINO or CLIP similarity, which mainly reflect semantic or identity consistency, these metrics directly measure whether the generated output matches the desired geometry and layout under our evaluation protocol.

\noindent\textbf{Implementation protocol.}
Unless otherwise stated, the retrieval stage recalls the top-$M=20$ candidates, and the selector uses a reference budget of $K=3$ for the final generator input. The default LogDet hyperparameters are $\gamma=3$ and $\tau=0.5$. All retrieval strategies compared in Table~\ref{tab:generation} use the same fine-tuned Flux.2 generator so that the study isolates the value of view-aware conditioning.

\subsection{Evaluation of Cross-instance Viewpoint Retrieval}

We first evaluate whether the proposed embedding actually learns cross-instance viewpoint comparability. Each test case contains a query view and a candidate set from a different object, annotated with relative viewpoint difference to the query. We rank candidates by predicted similarity and average the ranking metrics across queries.

\noindent\textbf{Main retrieval results.}
Table~\ref{tab:viewpoint_retrieval} shows a clear gap between generic semantic embeddings and retrieval models trained specifically for viewpoint. Generic backbones perform poorly on this benchmark, with NDCG@1 values between 0.329 and 0.359 and near-zero Spearman correlation. This confirms that semantic similarity is a poor proxy for cross-instance viewpoint compatibility. Task-specific fine-tuning improves all backbones substantially.

Our retriever performs best on every reported metric, reaching 0.750 NDCG@1, 0.859 NDCG@10, 0.832 Pairwise Concordance, and 0.710 Spearman's $\rho$. Relative to the strongest fine-tuned baseline Qwen3-VL$\dagger$, the gains are +0.064 on NDCG@1 and +0.052 on Spearman's $\rho$. Because the compared methods share strong pre-trained visual backbones, this gap is more consistent with the design of the retrieval head and supervision objective than with backbone capacity alone.

\noindent\textbf{Why the retriever matters.}
These retrieval results matter beyond ranking accuracy: if the generator is conditioned on geometrically mismatched references, the desired viewpoint evidence is already missing before denoising begins.

\subsection{Cross-subject View-aligned Generation}

We next evaluate the complete RAVA pipeline on 1k held-out object pairs. For each pair, the model receives one anchor image of subject $A$ and a candidate pool for subject $B$, and must generate subject $B$ at the anchor viewpoint under a multi-reference evaluation protocol.

\noindent\textbf{Baselines.}
Table~\ref{tab:generation} compares two groups of methods. \emph{Zero-shot Baselines} test representative multi-subject or multi-image generators directly, using one anchor image of subject $A$ and one reference image of subject $B$ without task-specific adaptation. \emph{Retrieval Strategies} fix the same fine-tuned Flux.2 backbone and vary only the method used to choose the reference images from the target-subject candidate pool, while conditioning the generator on one anchor image of subject $A$ together with multiple reference images of subject $B$. This controlled setting isolates the effect of conditioning quality from the effect of the generator itself. Figure~\ref{fig:teaser}, by contrast, uses NanoBanana2 only as a motivating stress test: because it is closed-source, it cannot be evaluated under the same reproducible multi-reference protocol used in Table~\ref{tab:generation}.

\begin{figure*}[t]
\centering
\includegraphics[width=\textwidth]{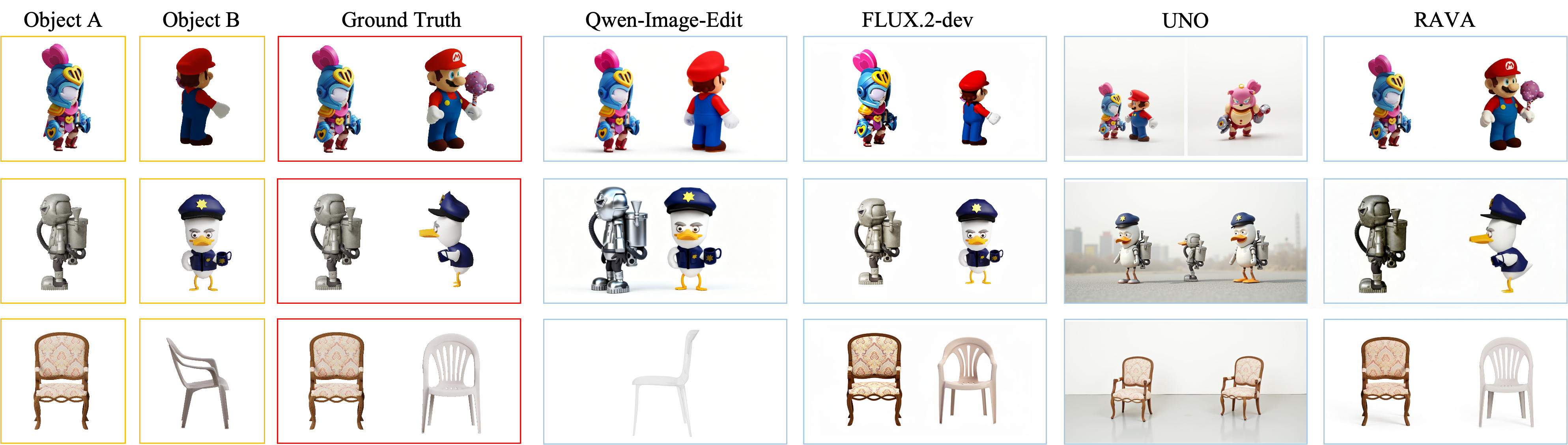}
\caption{Qualitative comparison for cross-subject view-aligned generation. Each row shows the anchor image of subject $A$, one input view of subject $B$, the ground-truth aligned target, three representative baselines, and the RAVA result. RAVA better preserves the anchor viewpoint while retaining target-specific structure.}
\Description{A grid of qualitative comparisons for cross-subject view-aligned generation. Each row contains an anchor image, a target-subject input image, a ground-truth aligned target view, several baseline generations, and the RAVA output in the rightmost column. The RAVA results better match the desired viewpoint and target-subject structure.}
\label{fig:visualization}
\end{figure*}

\noindent\textbf{Main generation results.}
The zero-shot baselines perform poorly on this task, even when they are strong general-purpose generators. Flux.2 is the best zero-shot baseline with PSNR 13.22, SSIM 0.7920, and LPIPS 0.3042, but still lags well behind retrieval-based conditioning. UNO is particularly weak among the zero-shot models, suggesting that the task stresses not only geometric transfer but also relational instruction following: the model must preserve one subject while applying its viewpoint to another. Overall, these results indicate that cross-subject viewpoint transfer is not handled reliably by generic image prompting alone.

Once Flux.2 is fine-tuned with multiple references, even random selection improves performance to 15.01 PSNR, 0.8249 SSIM, and 0.2143 LPIPS, showing that additional target-subject views provide useful geometric evidence. However, not all references are equally useful. Semantic retrievers such as CLIP and DINOv3 bring only marginal gains over random selection, which is consistent with their weak viewpoint retrieval behavior. Naive Top-K based on our viewpoint scores performs better, but RAVA achieves the strongest results overall. This ordering is aligned with the proposed decomposition: better viewpoint recall helps, and complementarity-aware selection helps further.

Two comparisons are especially informative. First, relative to the strongest zero-shot baseline Flux.2, RAVA improves PSNR by 2.58, increases SSIM by 0.0478, and reduces LPIPS by 0.1213. Second, relative to Top-K under the same Flux.2-FT backbone, RAVA gains +0.40 PSNR, +0.0090 SSIM, and -0.0180 LPIPS. The second comparison isolates the effect of subset selection under a fixed generator and a shared viewpoint retriever, showing that simply retrieving similar views is not yet sufficient.

\noindent\textbf{Qualitative comparison.}
Figure~\ref{fig:visualization} complements Table~\ref{tab:generation}. The baseline columns frequently preserve coarse subject identity yet still miss the anchor viewpoint, collapse thin structures, or omit target-specific components. UNO is notably brittle in this figure: beyond geometric errors, it often fails to respect the relational instruction implied by the prompt. By contrast, the rightmost RAVA column more consistently matches the desired viewing direction while preserving the geometry of subject $B$. The qualitative cases therefore reflect the same trend as the quantitative metrics rather than introducing a separate claim. They also suggest that the main gain of RAVA lies in structural alignment of the target subject, not merely in texture cleanup or superficial realism improvements. This visual tendency is also consistent with the metric gains reported in Table~\ref{tab:generation}.

\begin{table}[t]
  \centering
  \caption{Ablation study of feature fusion strategies for multi-granularity pooling. Best results are in \textbf{bold}.}
  \label{tab:ablation_fusion}
  \setlength{\tabcolsep}{8pt}
  \begin{tabular}{l *{3}{S[table-format=1.3]}}
    \toprule
    Fusion Strategy & {NDCG@1} & {Concord.} & {Spearman} \\
    \midrule
    Concat      & 0.712 & 0.795 & 0.668 \\
    Attention   & 0.731 & 0.814 & 0.692 \\
    Gated       & {\bfseries 0.750} & {\bfseries 0.832} & {\bfseries 0.710} \\
    \bottomrule
  \end{tabular}
\end{table}

\begin{table}[t]
  \centering
  \caption{Ablation study of hyperparameters $\gamma$ and $\tau$ for LogDet-based reference selection. $\gamma$ controls the viewpoint-alignment bias, and $\tau$ sets the redundancy-suppression temperature. The default configuration ($\gamma=3$, $\tau=0.5$) performs best.}
  \label{tab:ablation_hyperparams}
  \setlength{\tabcolsep}{10pt}
  \begin{tabular}{cc ccc}
    \toprule
    $\gamma$ & $\tau$ & PSNR $\uparrow$ & SSIM $\uparrow$ & LPIPS $\downarrow$ \\
    \midrule
    1 & 0.5 & 15.42 & 0.8211 & 0.1985 \\
    5 & 0.5 & 15.58 & 0.8312 & 0.1910 \\
    3 & 0.1 & 15.51 & 0.8285 & 0.1944 \\
    3 & 1.0 & 15.63 & 0.8340 & 0.1892 \\
    \midrule
    3 & 0.5 & \textbf{15.80} & \textbf{0.8398} & \textbf{0.1829} \\
    \bottomrule
  \end{tabular}
\end{table}

\subsection{Ablation Studies and Analysis}

\noindent\textbf{Fusion strategy.}
Table~\ref{tab:ablation_fusion} studies how the two feature granularities are combined. Direct concatenation already yields a strong retriever, but adaptive fusion is consistently better. Compared with concatenation, gated fusion improves NDCG@1 from 0.712 to 0.750 and Spearman's $\rho$ from 0.668 to 0.710. This supports the intuition that different objects require different balances of global and regional viewpoint evidence.

\noindent\textbf{Selection hyperparameters.}
Table~\ref{tab:ablation_hyperparams} analyzes the two selection hyperparameters. Reducing $\gamma$ to 1 weakens viewpoint bias and degrades all metrics, while increasing it to 5 improves over the weak-bias setting but remains worse than the default. Similarly, either overly sharp redundancy suppression ($\tau=0.1$) or overly soft suppression ($\tau=1.0$) underperforms the default $\tau=0.5$. These trends support the intended trade-off: the selector should favor well-aligned references, but not so aggressively that it collapses onto near-duplicates.

\noindent\textbf{Reference budget.}
Figure~\ref{fig:ablation_k} shows that increasing the number of references is not always beneficial. Performance improves when moving from a single reference to a small multi-reference set, peaks around $K=3$, and then degrades as additional references are introduced. Inference time, by contrast, increases monotonically. This pattern supports the practical motivation for selection: under a fixed conditioning budget, more references are useful only when they add non-redundant evidence.

\noindent\textbf{Discussion.}
Taken together, the experiments support a consistent interpretation: generic embeddings are weak viewpoint retrievers, viewpoint-aware retrieval improves the relevance of the candidate pool, and LogDet selection further improves the conditioning set by suppressing redundancy. This progression also matches the structure of the method itself. Each stage removes a specific source of ambiguity before generation, so the downstream gains are not only empirical but also mechanistically aligned with the intended design.

\subsection{Limitations}

Our study is conducted on rendered Objaverse-style objects, which is suitable for controlled viewpoint evaluation but does not remove the challenges of real-image deployment. In addition, RAVA assumes that the target-subject candidate pool already contains at least moderately relevant viewpoints; when the pool is very sparse or biased, retrieval becomes the dominant bottleneck.

\section{Conclusion}

We presented RAVA, a retrieval-augmented framework for cross-subject viewpoint alignment in reference-driven image generation. By retrieving viewpoint-compatible target-subject images and then selecting a compact complementary subset, RAVA provides the generator with stronger geometric evidence before synthesis. Across both retrieval and generation benchmarks, the results show that explicit geometric conditioning is substantially more effective than relying on end-to-end multi-image generation alone.

\bibliographystyle{ACM-Reference-Format}
\bibliography{sample-base}

\end{document}